\documentclass[runningheads]{llncs}
\usepackage{graphicx}
\usepackage{subfigure}
\usepackage{amsmath,amssymb} % define this before the line numbering.
\usepackage{color}
\usepackage{makecell}
\usepackage{bbding}
\usepackage{multirow}
\usepackage{threeparttable}
\usepackage[percent]{overpic}
\usepackage{hyperref}
\begin{document}
\title{Receptive Field Block Net for Accurate and Fast Object Detection}
% Replace with your title

\titlerunning{RFB Net for Accurate and Fast Object Detection}
% Replace with a meaningful short version of your title
%
\author{Songtao Liu, Di Huang\thanks{indicates corresponding author (ORCID: 0000-0002-2412-9330).}, and
Yunhong Wang}

%
%Please write out author names in full in the paper, i.e. full given and family names.
%If any authors have names that can be parsed into FirstName LastName in multiple ways, please include the correct parsing, in a comment to the volume editors:
%\index{Lastnames, Firstnames}
%(Do not uncomment it, because you may introduce extra index items if you do that, we will use scripts for introducing index entries...)
\authorrunning{Songtao Liu, Di Huang, and Yunhong Wang}
% Replace with shorter version of the author list. If there are more authors than fits a line, please use A. Author et al.
%
\institute{Beijing Advanced Innovation Center for Big Data and Brain Computing\\
Beihang University, Beijing 100191, China\\
\email{\{liusongtao, dhuang, yhwang\}@buaa.edu.cn}}

\maketitle              % typeset the header of the contribution

\begin{abstract}
Current top-performing object detectors depend on deep CNN backbones, such as ResNet-101 and Inception, benefiting from their powerful feature representations but suffering from high computational costs. Conversely, some lightweight model based detectors fulfil real time processing, while their accuracies are often criticized. In this paper, we explore an alternative to build a fast and accurate detector by strengthening lightweight features using a hand-crafted mechanism. Inspired by the structure of Receptive Fields (RFs) in human visual systems, we propose a novel RF Block (RFB) module, which takes the relationship between the size and eccentricity of RFs into account, to enhance the feature discriminability and robustness. We further assemble RFB to the top of SSD, constructing the RFB Net detector. To evaluate its effectiveness, experiments are conducted on two major benchmarks and the results show that RFB Net is able to reach the performance of advanced very deep detectors while keeping the real-time speed. Code is available at \url{https://github.com/ruinmessi/RFBNet}.
\keywords{Real-time Object Detection; Receptive Field Block (RFB) }
\end{abstract}

\section{Introduction}
\begin{figure}[t]
\begin{center}
   \includegraphics[width=0.75\linewidth]{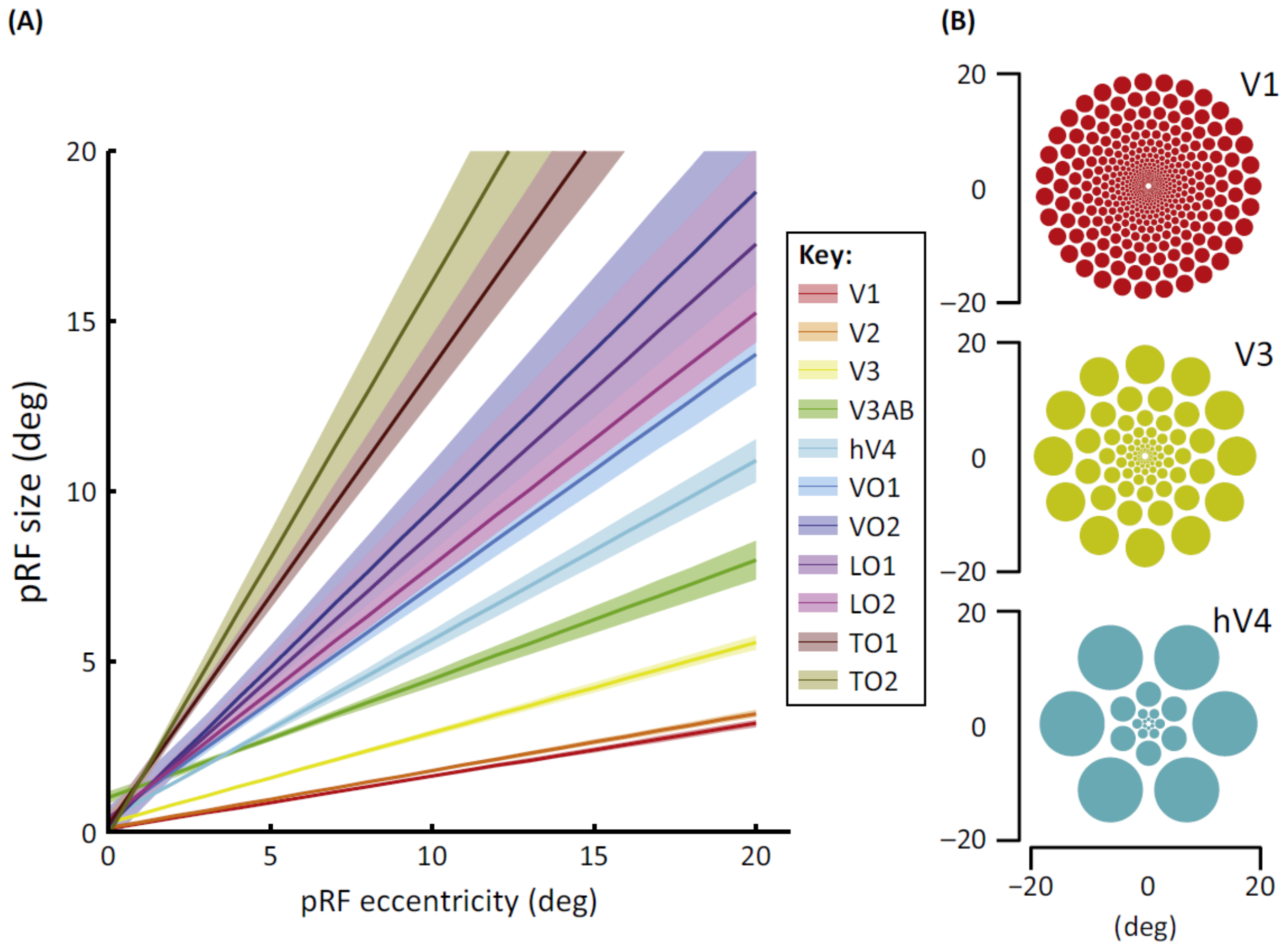}
\end{center}
   \caption{Regularities in human population Receptive Field (pRF) properties. \textbf{(A)} pRF size as a function of eccentricity in some human retinotopic maps, where two trends are evident: (1) the pRF size increases with eccentricity in each map and (2) the pRF size differs between maps. \textbf{(B)} The spatial array of the pRFs based on the parameters in \textbf{(A)}. The radius of each circle is the apparent RF size at the appropriate eccentricity. Reproduced from \cite{pRF-1} with the permission from J. Winawer and H. Horiguchi ({\url{https://archive.nyu.edu/handle/2451/33887}}).}
\label{fig:RF}
\end{figure}

In recent years, Region-based Convolutional Neural Networks (R-CNN) \cite{RCNN}, along with its representative updated descendants, \emph{e.g.} Fast R-CNN \cite{fastCNN} and Faster R-CNN \cite{FasterCNN}, have persistently promoted the performance of object detection on major challenges and benchmarks, such as Pascal VOC \cite{Pascal-voc}, MS COCO \cite{MS-COCO}, and ILSVRC \cite{imagenet}. They formulate this issue as a two-stage problem and build a  typical pipeline, where the first phase hypothesizes category-agnostic object proposals within the given image and the second phase classifies each proposal according to CNN based deep features. It is generally accepted that in these methods, CNN representation plays a crucial role, and the learned feature is expected to deliver a high discriminative power encoding object characteristics and a good robustness especially to moderate positional shifts (usually incurred by inaccurate boxes). A number of very recent efforts have confirmed such a fact. For instance, \cite{ResNet} and \cite{trade-off} extract features from deeper CNN backbones, like ResNet \cite{ResNet} and Inception \cite{inceptionv4}; \cite{FPN} introduces a top-down architecture to construct feature pyramids, integrating low-level and high-level information; and the latest top-performing Mask R-CNN \cite{mask-rcnn} produces an RoIAlign layer to generate more precise regional features. All these methods adopt improved features to reach better results; however, such features basically come from deeper neural networks with heavy computational costs, making them suffer from a low inference speed.

To accelerate detection, a single-stage framework is investigated, where the phase of object proposal generation is discarded. Although the pioneering attempts, namely You Look Only Once (YOLO) \cite{yolo} and Single Shot Detector (SSD) \cite{ssd}, illustrate the ability of real-time processing, they tend to sacrifice accuracies, with a clear drop ranging from 10\% to 40\% relative to state-of-the-art two-stage solutions \cite{focal-loss}. More recently, Deconvolutional SSD (DSSD) \cite{dssd} and RetinaNet \cite{focal-loss} substantially ameliorate the precision scores, which are comparable to the top ones reported by the two-stage detectors. Unfortunately their performance gains are credited to the very deep ResNet-101 \cite{ResNet} model as well, which limits the efficiency.

According to the discussion above, to build a fast yet powerful detector, a reasonable alternative is to enhance feature representation of the lightweight network by bringing in certain hand-crafted mechanisms rather than stubbornly deepening the model. On the other side, several discoveries in neuroscience reveal that in human visual cortex, the size of population Receptive Field (pRF) is a function of eccentricity in their retinotopic maps, and although varying between maps, it increases with eccentricity in each map \cite{pRF-1}, as illustrated in Fig. \ref{fig:RF}. It helps to highlight the importance of the region nearer to the center and elevate the insensitivity to small spatial shifts. A few shallow descriptors coincidentally make use of this mechanism to design \cite{daisy,huang2014hsog,weng2015derf} or learn \cite{discriminative,learninglocal,learningconvex} their pooling schemes, and show good performance in matching image patches.

Regarding current deep learning models, they commonly set RFs at the same size with a regular sampling grid on a feature map, which probably induces some loss in the feature discriminability as well as robustness. Inception \cite{inceptionv1} considers RFs of multiple sizes, and it implements this concept by launching multi-branch CNNs with different convolution kernels. Its variants \cite{inceptionv2,inceptionv4,pvanet} achieve competitive results in object detection (in the two-stage framework) and classification tasks. However, all kernels in Inception are sampled at the same center. A similar idea appears in \cite{deeplabv3}, where an Atrous Spatial Pyramid Pooling (ASPP) is exploited to capture multi-scale information. It applies several parallel convolutions with different atrous rates on the top feature map to vary the sampling distance from the center, which proves effective in semantic segmentation. But the features only have a uniform resolution from previous convolution layers of the same kernel size, and compared to the daisy shaped ones, the resulting feature tends to be less distinctive. Deformable CNN \cite{deformableCNN} attempts to adaptively adjust the spatial distribution of RFs according to the scale and shape of the object. Although its sampling grid is flexible, the impact of eccentricity of RFs is not taken into account, where all pixels in an RF contribute equally to the output response and the most important information is not emphasized.

\begin{figure*}[t]
\begin{center}
   \includegraphics[width=0.95\linewidth]{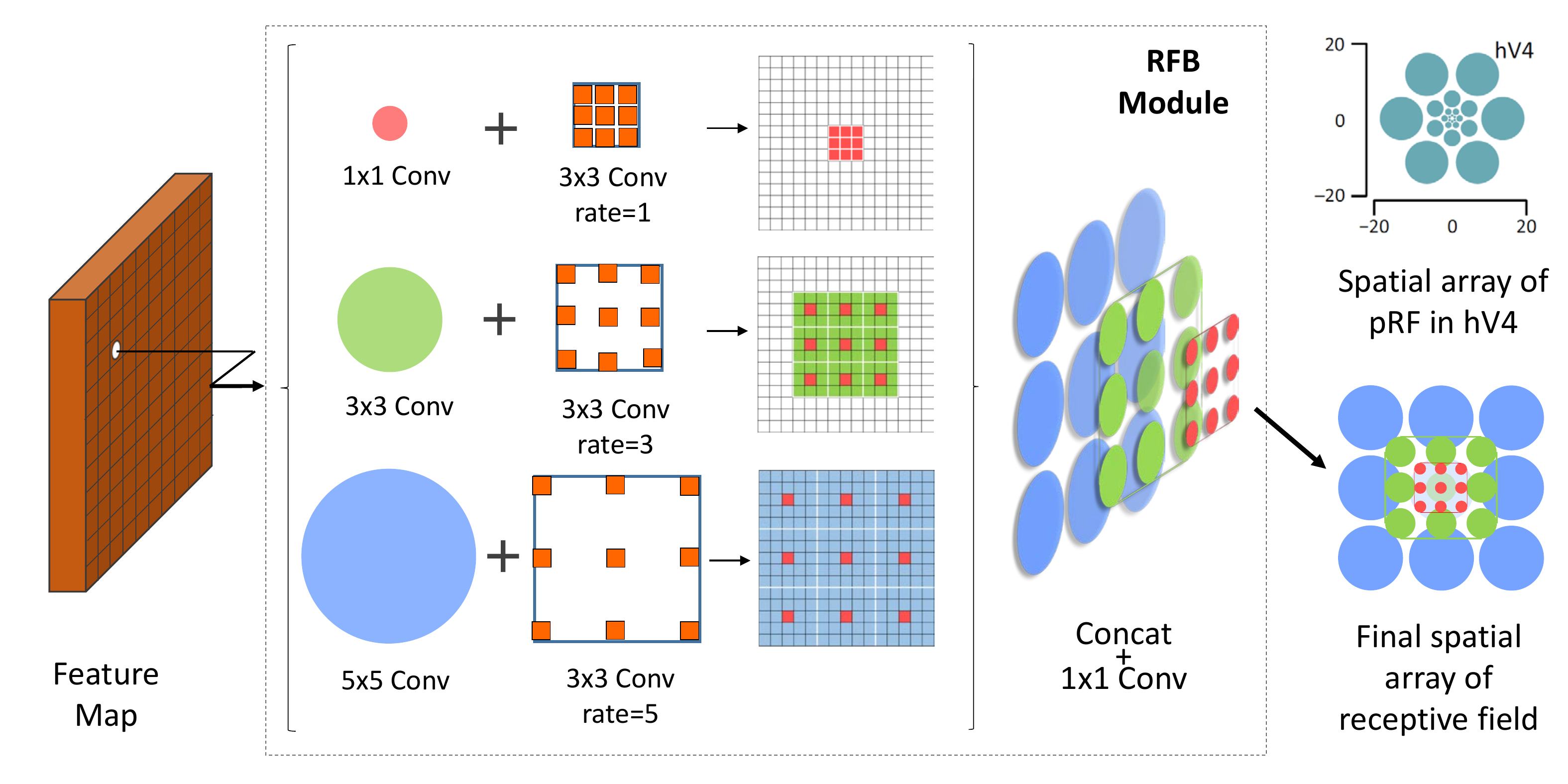}
\end{center}
   \caption{Construction of the RFB module by combining multiple branches with different kernels and dilated convolution layers. Multiple kernels are analogous to the pRFs of varying sizes, while dilated convolution layers assign each branch with an individual eccentricity to simulate the ratio between the size and eccentricity of the pRF. With a concatenation and 1$\times $1 conv in all the branches, the final spatial array of RF is produced, which is similar to that in human visual systems, as depicted in Fig.~\ref{fig:RF}.}
\label{fig:RFB}
\end{figure*}

Inspired by the structure of RFs in the human visual system, this paper proposes a novel module, namely Receptive Field Block (RFB), to strengthen the deep features learned from lightweight CNN models so that they can contribute to fast and accurate detectors. Specifically, RFB makes use of multi-branch pooling with varying kernels corresponding to RFs of different sizes, applies dilated convolution layers to control their eccentricities, and reshapes them to generate final representation, as in Fig. \ref{fig:RFB}. We then assemble the RFB module to the top of SSD \cite{ssd}, a real-time approach with a lightweight backbone, and construct an advanced one-stage detector (RFB Net). Thanks to such a simple module, RFB Net delivers relatively decent scores that are comparable to the ones of up-to-date deeper backbone network based detectors \cite{FPN,FCIS,focal-loss} and retains the fast speed of the original lightweight detector. Additionally, the RFB module is generic and imposes few constraints on the network architecture.

Our main contributions can be summarized as follows:
\begin{enumerate}
\setlength{\itemsep}{3pt}
\setlength{\parsep}{0pt}
\setlength{\parskip}{0pt}
  \item We propose the RFB module to simulate the configuration in terms of the size and eccentricity of RFs in human visual systems, aiming to enhance deep features of lightweight CNN networks.
  \item We present the RFB Net based detector, and by simply replacing the top convolution layers of SSD \cite{ssd} with RFB, it shows significant performance gain while still keeping the computational cost under control.
  \item We show that RFB Net achieves state-of-the-art results on the Pascal VOC and MS COCO at a real time processing speed, and demonstrate the generalization ability of RFB by linking it to MobileNet \cite{mobilenets}.

\end{enumerate}

\section{Related Work}

\textbf{Two-stage detector:} R-CNN \cite{RCNN} straightforwardly combines the steps of cropping box proposals like Selective Search \cite{selective-search} and classifying them through a CNN model, yielding a significant accuracy gain compared to traditional methods, which opens the deep learning era in object detection.
Its descendants (\emph{e.g.}, Fast R-CNN \cite{fastCNN}, Faster R-CNN \cite{FasterCNN}) update the two-stage framework and achieve dominant performance. Besides, a number of effective extensions are proposed to further improve the detection accuracy, such as R-FCN \cite{R-FCN}, FPN \cite{FPN}, Mask R-CNN \cite{mask-rcnn}.

%-------------------------------------------------------------------------
\textbf{One-stage detector:} The most representative one-stage detectors are YOLO \cite{yolo,yolo9000} and SSD \cite{ssd}. They predict confidences and locations for multiple objects based on the whole feature map. Both the detectors adopt lightweight backbones for acceleration, while their accuracies apparently trail those of top two-stage methods.

Recent more advanced single-stage detectors (\emph{e.g.}, DSSD \cite{dssd} and RetinaNet \cite{focal-loss}) update their original lightweight backbones by the deeper ResNet-101 and apply certain techniques, such as deconvolution \cite{dssd} or Focal Loss \cite{focal-loss}, whose scores are comparable and even superior to the ones of state-of-the-art two-stage methods. However, such performance gains largely consume their advantage in speed.

\textbf{Receptive field:} Recall that in this study, we aim to improve the performance of high-speed single-stage detectors without incurring too much computational burden. Therefore, instead of applying very deep backbones, RFB, imitating the mechanism of RFs in the human visual system, is used to enhance lightweight model based feature representation. Actually, there exist several studies that discuss RFs in CNN, and the most related ones are the Inception family \cite{inceptionv1,inceptionv2,inceptionv4}, ASPP \cite{deeplabv3}, and Deformable CNN \cite{deformableCNN}.

\begin{figure*}[t]
\begin{center}
   \includegraphics[width=1.0\linewidth]{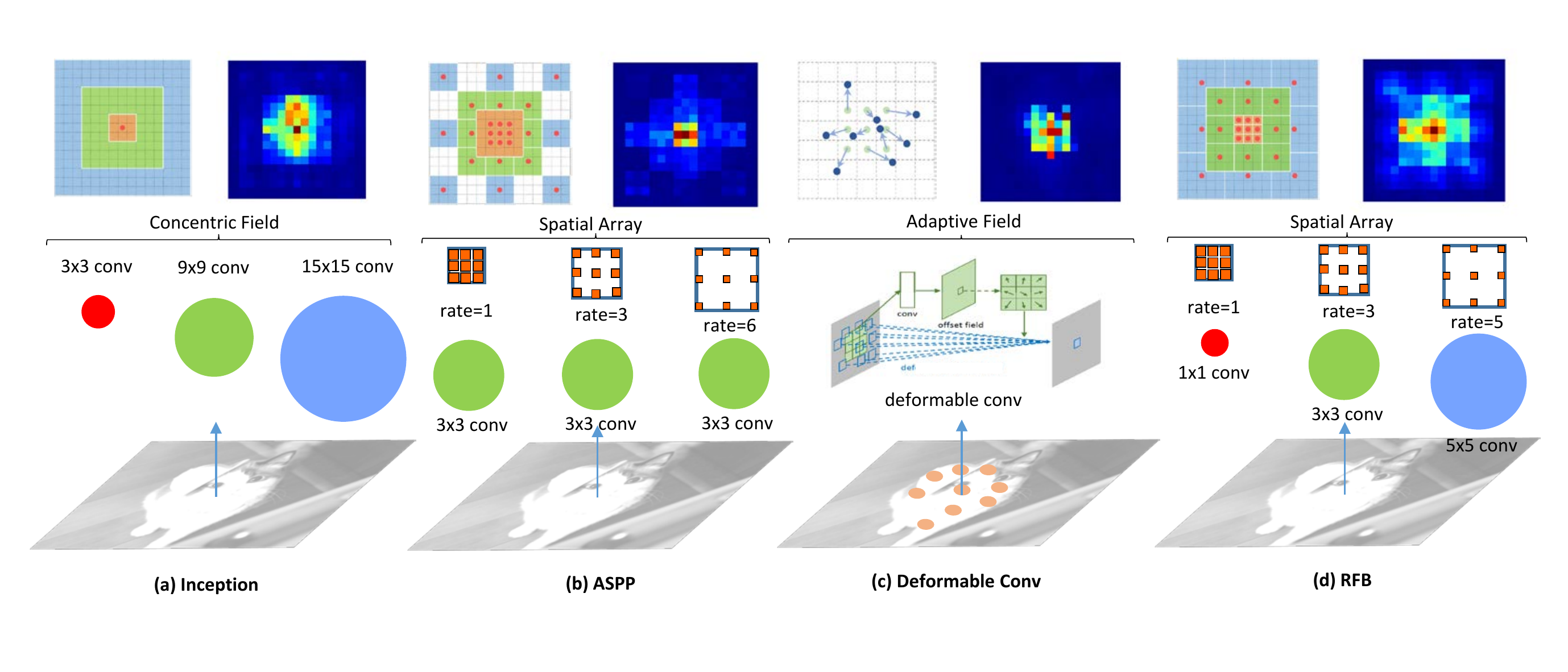}
\end{center}
   \caption{Four typical structures of Spatial RFs. (a) shows the kernels of multiple sizes in Inception. (b) demonstrates the daisy-like pooling configuration in ASPP. (c) adopts deformable conv to produce an adaptive RF according to object characteristics. (d) illustrates the mechanism of RFB. The color map of each structure is the effective RF derived from one correspondent layer in the trained model, depicted by the same gradient back-propagation method in \cite{ERF}.  In (a) and (b), we adjust the RF sizes in original Inception and ASPP for fair comparison.}
\label{fig:compare}
\end{figure*}

The Inception block adopts multiple branches with different kernel sizes to capture multi-scale information. However, all the kernels are sampled at the same center, which requires much larger ones to reach the same sampling coverage and thus loses some crucial details. For ASPP, dilated convolution varies the sampling distance from the center, but the features have a uniform resolution from the previous convolution layers of the same kernel size, which treats the clues at all the positions equally, probably leading to confusion between object and context. Deformable CNN \cite{deformableCNN} learns distinctive resolutions of individual objects, unfortunately it holds the same downside as ASPP. RFB is indeed different from them, and it highlights the relationship between RF size and eccentricity in a daisy-shape configuration, where bigger weights are assigned to the positions nearer to the center by smaller kernels, claiming that they are more important than the farther ones. See Fig.~\ref{fig:compare} for differences of the four typical spatial RF structures. On the other side, Inception and ASPP have not been successfully adopted to improve one-stage detectors, while RFB shows an effective way to make use of their advantages in this issue.

%-------------------------------------------------------------------------
%-------------------------------------------------------------------------
%-------------------------------------------------------------------------
\section{Method}
In this section, we revisit the human visual cortex, introduce our RFB components and the way to simulate such a mechanism, and describe the architecture of the RFB Net detector as well as its training/testing schedule.

\subsection{Visual Cortex Revisit}
During the past few decades, it has come true that functional Magnetic Resonance Imaging (fMRI) non-invasively measures human brain activities at a resolution in millimeter, and RF modeling has become an important sensory science tool used to predict responses and clarify brain computations. Since human neuroscience instruments often observe the pooled responses of many neurons, these models are thus commonly called pRF models \cite{pRF-1}. Based on fMRI and pRF modeling, it is possible to investigate the relation across many visual field maps in the cortex. At each cortical map, researchers find a positive correlation between pRF size and eccentricity \cite{pRF-1}, while the coefficient of correlation varies in visual field maps, as shown in Fig.~\ref{fig:RF}.

%-------------------------------------------------------------------------
\subsection{Receptive Field Block}

The proposed RFB is a multi-branch convolutional block. Its inner structure can be divided into two components: the multi-branch convolution layer with different kernels and the trailing dilated pooling or convolution layers. The former part is identical to that of Inception, responsible for simulating the pRFs of multiple sizes, and the latter part reproduces the relation between the pRF size and eccentricity in the human visual system. Fig.~\ref{fig:RFB} illustrates RFB along with its corresponding spatial pooling region maps. We elaborate the two parts and their functions in detail in the following.

\textbf{Multi-branch convolution layer:} According to the definition of RF in CNNs, it is a simple and natural way to apply different kernels to achieve multi-size RFs, which is supposed to be superior to the RFs that share a fixed size.

%The Inception series \cite{inceptionv1,inceptionv2,inceptionv4} clearly demonstrate the effectiveness of this construction in object detection and image classification \cite{trade-off}.

We adopt the latest changes in the updated versions, \emph{i.e.}, Inception V4 and Inception-ResNet V2 \cite{inceptionv4} in the Inception family. To be specific, first, we employ the bottleneck structure in each branch, consisting of a $1\times 1$ conv-layer, to decrease the number of channels in the feature map plus an $n\times n$ conv-layer. Second, we replace the $5\times 5$ conv-layer by two stacked $3\times 3$ conv-layers to reduce parameters and deeper non-linear layers. For the same reason, we use a $1\times n$ plus an $n\times 1$ conv-layer to take place of the original $n\times n$ conv-layer. Ultimately, we apply the shortcut design from ResNet \cite{ResNet} and Inception-ResNet V2 \cite{inceptionv4}.

%Since the top convolution layers often work with a stride of 2 or decreased feature maps in output, we change the shortcut from identity mapping to a $1\times 1$ conv-layer without non-linear activation.

\textbf{Dilated pooling or convolution layer:} This concept is originally introduced in Deeplab \cite{deeplab}, which is also named the astrous convolution layer. The basic intention of this structure is to generate feature maps of a higher resolution, capturing information at a larger area with more context while keeping the same number of parameters. This design has rapidly proved competent at semantic segmentation \cite{deeplabv3}, and has also been adopted in some reputable object detectors, such as SSD \cite{ssd} and R-FCN \cite{R-FCN}, to elevate speed or/and accuracy.

\begin{figure}[t]
\centering
\subfigure[RFB]{
\begin{minipage}[t]{0.45\textwidth}
\centering
\includegraphics[width=1.0\textwidth]{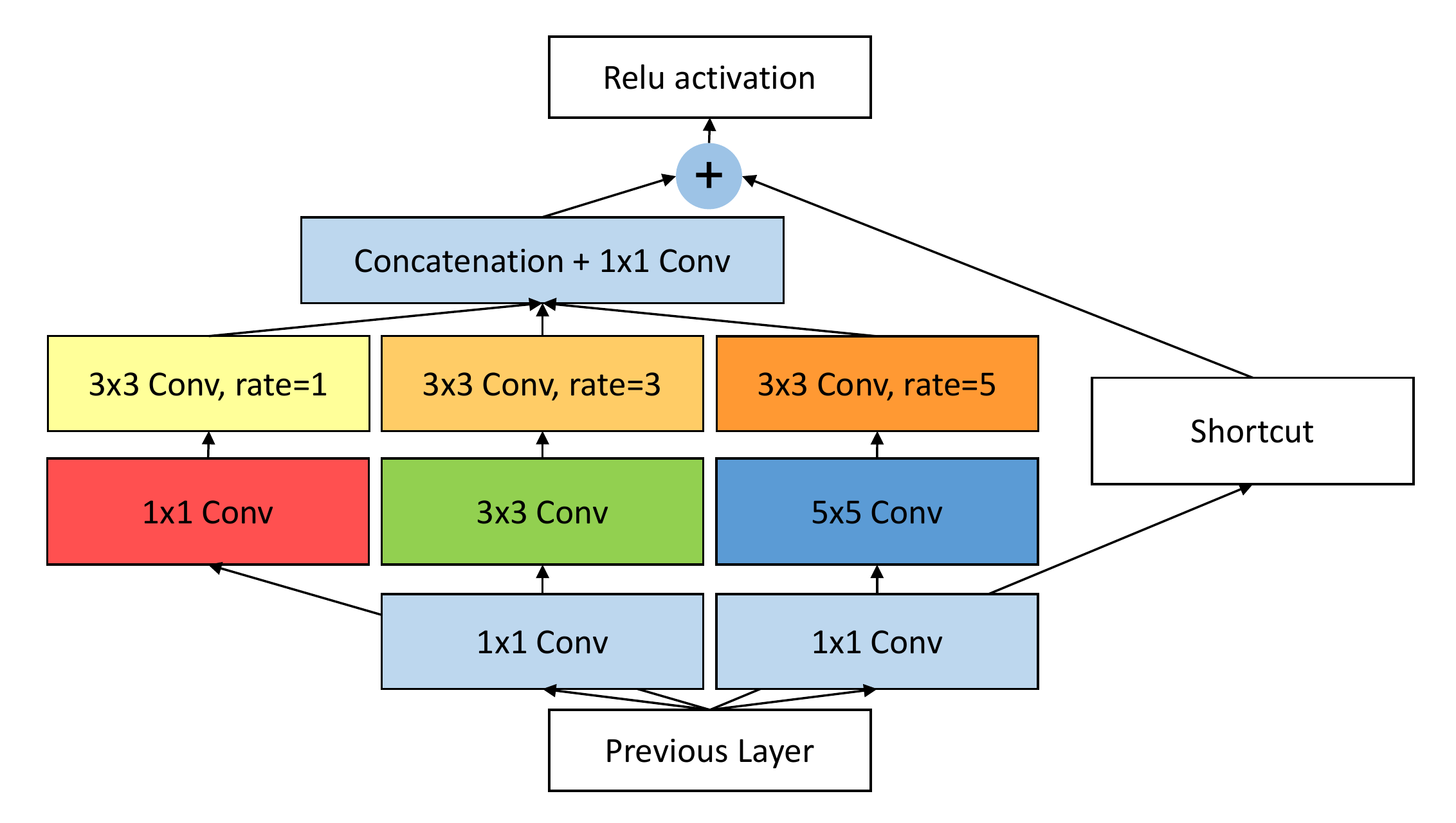}
\end{minipage}
}
\subfigure[RFB-s]{
\begin{minipage}[t]{0.45\textwidth}
\centering
\includegraphics[width=1.0\textwidth]{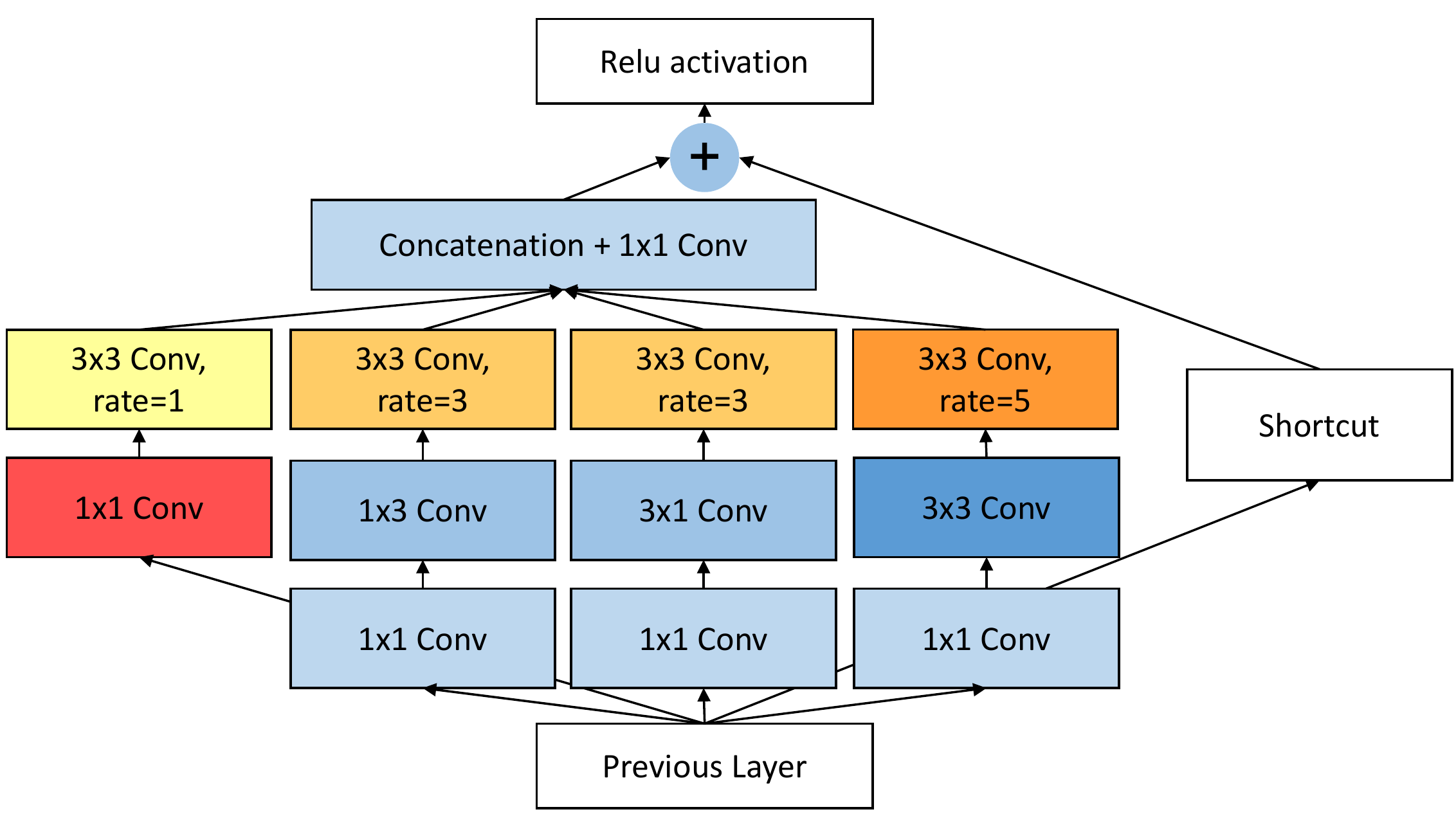}
\end{minipage}
}
\caption{The architectures of RFB and RFB-s. RFB-s is employed to mimic smaller pRFs in shallow human retinotopic maps, using more branches with smaller kernels. Following \cite{inceptionv2}, we use two layers of $3\times 3$ conv replacing $5\times 5$ to reduce parameters, which is not shown for better visualization.} %% label for entire figure
\label{fig:RFB-a&b}
\end{figure}

In this paper, we exploit dilated convolution to simulate the impact of the eccentricities of pRFs in the human visual cortex. Fig.~\ref{fig:RFB-a&b} illustrates two combinations of multi-branch convolution layer and dilated pooling or convolution layer. At each branch, the convolution layer of a particular kernel size is followed by a pooling or convolution layer with a corresponding dilation. The kernel size and dilation have a similar positive functional relation as that of the size and eccentricity of pRFs in the visual cortex. Eventually, the feature maps of all the branches are concatenated, merging into a spatial pooling or convolution array as in Fig.~\ref{fig:RF}.

%When using a max or an average pooling layer with a dilation, RFB contains less parameters, but it loses the flexibility in reorganizing the features from RFs of different sizes. While choosing a dilated convolution layer, RFB enhances the feature representation power since the final feature map can be seen as a learnable linear combination of the spatial array, although its number of parameters moderately increases (we show the trade-off in the selection step in Section~\ref{section:ablation}).

The specific parameters of RFB, \emph{e.g.}, kernel size, dilation of each branch, and number of branches, are slightly different at each position within the detector, which are clarified in the next section.

%-------------------------------------------------------------------------
\subsection{RFB Net Detection Architecture}
\label{section:arch}

\begin{figure*}[t]
\begin{center}
   \includegraphics[width=0.85\linewidth]{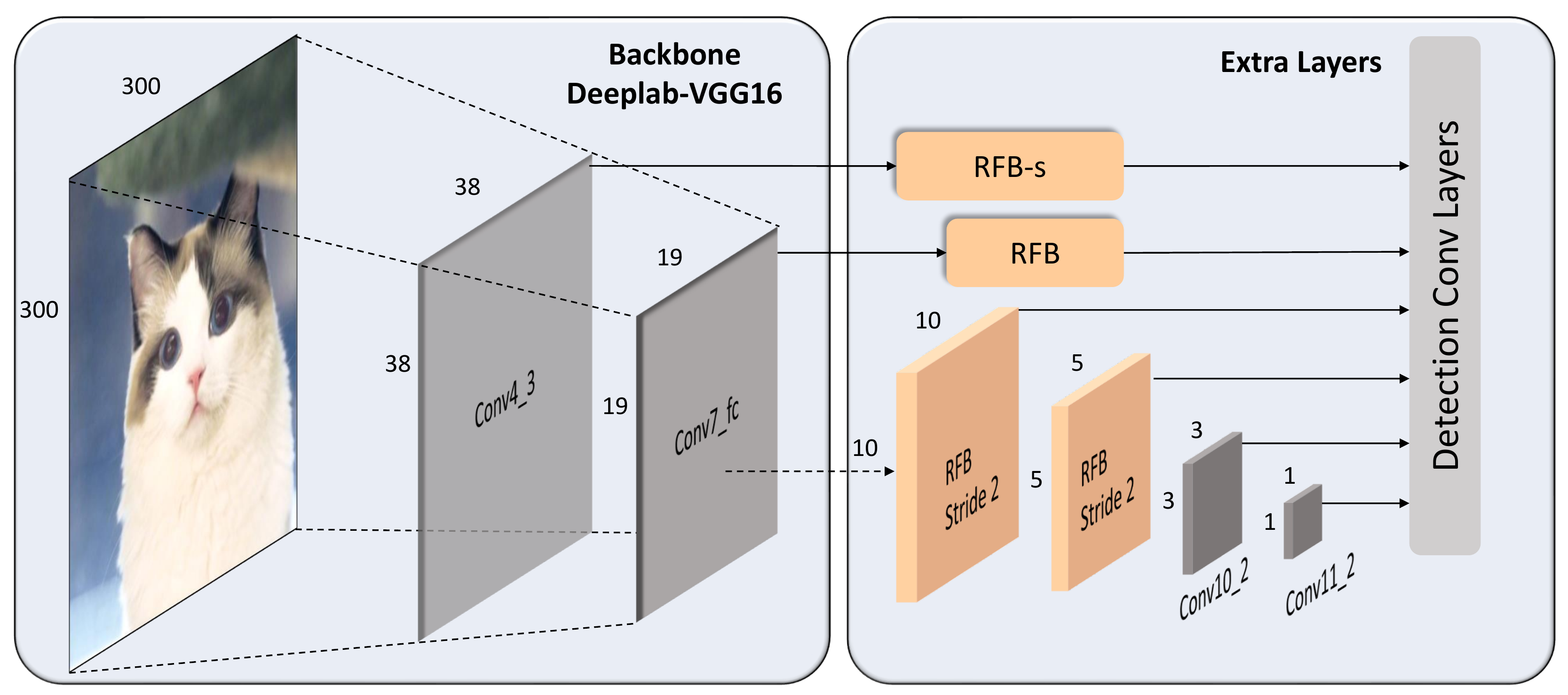}
\end{center}
   \caption{The pipeline of RFB-Net300. The conv4\_3 feature map is tailed by RFB-s which has smaller RFs and an RFB module with stride 2 is produced by operating 2-stride multi-kernel conv-layers in the original RFB.}
\label{fig:pipeline}
\end{figure*}

The proposed RFB Net detector reuses the multi-scale and one-stage framework of SSD \cite{ssd}, where the RFB module is embedded to ameliorate the feature extracted from the lightweight backbone so that the detector is more accurate and still fast enough. Thanks to the property of RFB for easily being integrated into CNNs, we can preserve the SSD architecture as much as possible. The main modification lies in replacing the top convolution layers with RFB, and some minor but active ones are given in Fig.~\ref{fig:pipeline}.

\textbf{Lightweight backbone:} We use exactly the same backbone network as in SSD \cite{ssd}. In brief, it is a VGG16 \cite{vgg} architecture pre-trained on the ILSVRC CLS-LOC dataset \cite{imagenet}, where its fc6 and fc7 layers are converted to convolutional layers with sub-sampling parameters, and its pool5 layer is changed from 2$\times $2-s2 to 3$\times $3-s1. The dilated convolution layer is used to fill ¡°holes¡± and all the dropout layers and the fc8 layer are removed. Even though many accomplished lightweight networks have recently been proposed (\emph{e.g.} DarkNet \cite{yolo9000}, MobileNet \cite{mobilenets}, and ShuffleNet \cite{shufflenet}), we focus on this backbone to achieve direct comparison to the original SSD \cite{ssd}.

\textbf{RFB on multi-scale feature maps:} In the original SSD \cite{ssd}, the base network is followed by a cascade of convolutional layers to form a series of feature maps with consecutively decreasing spatial resolutions and increasing fields of view. In our implementation, we keep the same cascade structure of SSD, but the front convolutional layers with feature maps of relatively large resolutions are replaced by the RFB module. In the primary version of RFB, we use a a single structure setting to imitate the impact of eccentricity. As the rate of the size and eccentricity of pRF differs between visual maps, we correspondingly adjust the parameters of RFB to form an RFB-s module, which mimics smaller pRFs in shallow human retinotopic maps, and put it behind the conv4\_3 features, as illustrated in Fig.~\ref{fig:RFB-a&b} and Fig.~\ref{fig:pipeline}. The last few convolutional layers are preserved since the resolutions of their feature maps are too small to apply filters with large kernels like $5\times 5$.

%-------------------------------------------------------------------------
\subsection{Training Settings}
We implement our RFB Net detector based on the framework of Pytorch\footnote{https://pytorch.org/}, utilizing several parts of open source infrastructures provided by the ssd.pytorch repository\footnote{https://github.com/amdegroot/ssd.pytorch}. Our training strategies mostly follow SSD, including data augmentation, hard negative mining, scale and aspect ratios for default boxes, and loss functions (\emph{e.g.}, smooth L1 loss for localization and softmax loss for classification), while we slightly change our learning rate scheduling for better accommodation of RFB. More details are given in the following section of experiments. All new conv-layers are initialized with the ¡°MSRA¡± method \cite{MSRA}.

%With a potentially large number of boxes generated from our detector, performing non-maximum suppression (NMS)  might significantly slow down the inference. We thus suggest a more efficient post-processing strategy when there are too many proposals. We first use a confidence threshold of 0.015 to filter out most boxes, which is a little higher than 0.01 in SSD, and we then pre-select the top 200 boxes with the largest scores and apply NMS with jaccard overlap of 0.45 for each class, leaving only 50 per class. Overall, final detection is conducted on the top 200 boxes per image. This step notably shrinks the computational time for NMS, especially in the COCO dataset with 80 classes, and has a marginal impact on accuracy.

\section{Experiments}
We conduct experiments on the Pascal VOC 2007 \cite{Pascal-voc} and MS COCO \cite{MS-COCO} datasets, which have 20 and 80 object categories respectively. In VOC 2007, a predicted bounding box is positive if its Intersection over Union (IoU) with the ground truth is higher than 0.5, while in COCO, it uses various thresholds for more comprehensive calculation. The metric to evaluate detection performance is the mean Average Precision (mAP).

%-------------------------------------------------------------------------
\subsection{Pascal VOC 2007}

In this experiment, we train our RFB Net on the union of 2007 \emph{trainval} set and 2012 \emph{trainval} set. We set the batch size at 32 and the initial learning rate at $10^{-3}$ as in the original SSD \cite{ssd}, but it makes the training process not so stable as the loss drastically fluctuates. Instead, we use a ``warmup'' strategy that gradually ramps up the learning rate from $10^{-6}$ to $4\times 10^{-3}$ at the first 5 epochs. After the ``warmup'' phase, it goes back to the original learning rate schedule, divided by 10 at 150 and 200 epochs. The total number of training epochs is 250. Following \cite{ssd}, we utilize a weight decay of 0.0005 and a momentum of 0.9.

Table~\ref{table:2007} shows the comparison between our results and the state of the art ones on the VOC2007 \emph{test} set. SSD300* and SSD512* are the updated SSD results with an expansion of data augmentation \cite{ssd}, which zooms out the images to create more small examples. For fair comparison, we reimplement SSD with Pytorch-0.3.0 and CUDNN V6, the same environment as that of RFB Net.  By integrating the RFB layers, our basic model, \emph{i.e.} RFB Net300, outperforms SSD and YOLO with an mAP of 80.5\%, while keeping the real-time speed as SSD300. It even reaches the same accuracy with R-FCN \cite{R-FCN}, the advanced model under the two-stage framework. RFB Net512 achieves the mAP of 82.2\% with a larger input size, better than most one stage and two stage object detection systems equipped with very deep base backbone networks, while still running at a high speed.

\begin{table}[t]
\begin{center}
\scalebox{.75}{
\resizebox{\textwidth}{!}{
\begin{threeparttable}
\begin{tabular}{l|l|l|l|l}
\multicolumn{1}{c|}{Method} & Backbone   & Data   & mAP(\%)  & FPS \\ \hline
Faster \cite{FasterCNN}                      & VGG        & 07+12 & 73.2 & 7   \\
Faster \cite{ResNet}                      & ResNet-101 & 07+12 & 76.4 & 5   \\
R-FCN \cite{R-FCN}                        & ResNet-101 & 07+12 & 80.5 & 9   \\
YOLOv2 544 \cite{yolo9000}                  & Darknet    & 07+12 & 78.6 & 40  \\
R-FCN w Deformable CNN \cite{deformableCNN}              & ResNet-101 & 07+12 & \textbf{82.6} & 8\tnote{\dag} \\ \hline
SSD300* \cite{ssd}                     & VGG        & 07+12 & 77.2 & \textbf{120}\tnote{\ddag}  \\
DSSD321 \cite{dssd}                     & ResNet-101 & 07+12 & 78.6 & 9.5 \\
\textbf{RFB Net300}             & VGG        & 07+12 & \textbf{80.5} &\textbf{83}  \\ \hline
SSD512* \cite{ssd}                     & VGG        & 07+12 & 79.8 & 50\tnote{\ddag}  \\
DSSD513 \cite{dssd}                     & ResNet-101 & 07+12 & 81.5 & 5.5 \\
\textbf{RFB Net512}            & VGG        & 07+12 & \textbf{82.2} & \textbf{38}
\end{tabular}
\begin{tablenotes}
    \footnotesize
    \item[\dag] {Extrapolated time}
    \item[\ddag] {Tested in Pytorch-0.3.0 and CUDNN V6 for fair comparison}
\end{tablenotes}
\end{threeparttable}}}
\end{center}
\caption{Comparison of detection methods on the PASCAL VOC 2007 \emph{test} set. All runtime information is computed on a Graphics card of Geforce GTX Titan X (Maxwell architecture).}
\label{table:2007}
\end{table}
%-------------------------------------------------------------------------
\subsection{Ablation Study}
\label{section:ablation}

\textbf{RFB module:} For better understanding RFB, we investigate the impact of each component in its design and compare RFB with some similar structures. The results are summarized in Table~\ref{table:ablation} and Table~\ref{table:arch}. As displayed in Table~\ref{table:ablation}, the original SSD300 with new data augmentation achieves a 77.2\% mAP. By simply replacing the last convolution layer with the RFB-max pooling, we can see that the result is improved to 79.1\%, delivering a gain of 1.9\%, which indicates that the RFB module is effective in detection.

\textbf{Cortex map simulation:} As described in Sec.\ref{section:arch}, we tune our RFB parameters to simulate the ratio between the size and eccentricity of pRFs in cortex maps. This adjustment boosts the performance by 0.5\% (from 79.1\% to 79.6\%) for RFB max pooling and 0.4\% for RFB dilated conv (from 80.1\% to 80.5\%), which validates the mechanism in human visual systems (Table~\ref{table:ablation}).

\textbf{More prior anchors:} The original SSD associates only 4 default boxes at conv4\_3, conv10\_2, and conv11\_2 feature map locations and 6 default anchors for all the other layers. Recent research \cite{tiny-face} claims that low level features are critical to detecting small objects. We thus suppose that performance, especially that of small instances, tends to increase if more anchors are added in low level feature maps like conv4\_3. In the experiment, we put 6 default priors at conv4\_3, and it has no influence on the original SSD, while it further improves 0.2\% (from 79.6\% to 79.8\%) for our RFB model (Table~\ref{table:ablation}).

\textbf{Dilated convolutional layer:} In early experiments, we choose dilated pooling layers for RFB to avoid incurring additional parameters, but these stationary pooling strategies limit feature fusion of RFs of multiple sizes. When picking the dilated convolutional layer, we find that it raises the accuracy by 0.7\% (from 79.8\% to 80.5\%) without slowing down the inference speed (Table~\ref{table:ablation}).
\begin{table}[thbp]
\begin{center}
\scalebox{.75}{
\resizebox{\textwidth}{!}{
\begin{tabular}{l|l|lllll|l}
                    & SSD* &            &            &            &             &            &\begin{tabular}[c]{@{}l@{}}RFB \\ \end{tabular}   \\ \hline
RFB-max pooling?     &      & \Checkmark & \Checkmark & \Checkmark &            &            \\
Add RFB-s?          &      &            & \Checkmark & \Checkmark & \Checkmark &            &\Checkmark \\
More Prior?          &      &            &            & \Checkmark & \Checkmark & \Checkmark &\Checkmark \\
RFB-avg pooling?     &      &            &            &            & \Checkmark &            \\
RFB-dilated conv?    &      &            &            &            &            & \Checkmark &\Checkmark \\ \hline
                     & 77.2 & 79.1       & 79.6       & 79.8       & 79.8       & 80.1        &\textbf{80.5}
\end{tabular}}}
\end{center}
\caption{Effectiveness of various designs on the VOC 2007 \emph{test} set (refer to Section~\ref{section:arch} and Section~\ref{section:ablation} for more details).}
\label{table:ablation}
\end{table}

%\textbf{Block architecture:} We also compare our RFB to Deformable CNN \cite{deformableCNN}, Inception \cite{inceptionv1}, Dilated Convolution \cite{deeplabv3}, ResNet \cite{ResNet}, ResNext \cite{resnext}, and several RFB-like modules with special settings. For the RFB-like modules, ``fixed eccentricity'' means all the dilated convolution layers at multi branches have the same dilation at 1; ``fixed RF'' sets the kernel size of all the convolution layers in the module to 3$\times $3; and ``negative ratio'' inverses the ratio of kernel size and dilation. We keep all the different structures have the same training schedule and almost the same number of parameters. Their evaluations are recorded in Table~\ref{table:arch}, and we can see that our RFB performs best.
%Specially, RFB surpasses Dilated Conv with rate at 8, which has the same size of entire RF. It points out that the dedicated RFB structure indeed contributes to the detection precision.

\textbf{Comparison with other architectures:} We also compare our RFB with Inception \cite{inceptionv1}, ASPP \cite{deeplabv3} and Deformable CNN \cite{deformableCNN}. For Inception, besides the original version, we change its parameters so that it has the same RF size as RFB does (termed ``Inception-L''). For ASPP, its primary parameters are tuned in image segmentation \cite{deeplabv3} and the RFs are too large for detection, and in our experiment, we set it at the same size as in RFB as well (termed ``ASPP-S''). Fig.~\ref{fig:compare} shows a visualized comparison in their structures. Simply, we individually mount these structures on the top layer of the detector as in Fig.~\ref{fig:pipeline} and keep the same training schedule and almost the same number of parameters. Their evaluations on the Pascal VOC and MS COCO  are recorded in Table~\ref{table:arch}, and we can see that our RFB performs best. It points out that the dedicated RFB structure indeed contributes to the detection precision, as it has a larger effective RF than the counterparts (see an example in Fig.~\ref{fig:compare}).

\begin{table}[thbp]
\begin{center}
\scalebox{.9}{
\resizebox{\textwidth}{!}{
\begin{tabular}{l|c|c|c}
Architecture                          & \#parameters & VOC 2007 mAP (\%) & COCO minival mAP (\%) \\ \hline
\textbf{RFB}                      & 34.5M        & \textbf{80.1}   & \textbf{29.7}                   \\ \hline
Inception \cite{inceptionv1}        & 32.9M        & 78.4              & 27.3                   \\
Inception-L                           & 33.3M        & 79.5              & 28.5                   \\
ASPP-S               & 33.4M        & 79.7              & 28.1                   \\
Deformable CNN \cite{deformableCNN} & 35.2M        & 79.5              & 27.6
\end{tabular}}}
\end{center}
\caption{Comparison of different blocks on VOC 2007 \emph{test} and MS COCO \emph{minival2014}.}
\label{table:arch}
\end{table}

%------------------------------------------------------------------------
\subsection{Microsoft COCO}

To further validate the proposed RFB module, we carry out experiments on the MS COCO dataset. Following \cite{ssd,focal-loss}, we use the \emph{trainval35k} set (\emph{train} set + \emph{val 35k} set) for training and set the batch size at 32. We keep the original SSD strategy that decreases the size of default boxes, since objects in COCO are smaller than those in PASCAL VOC. At the begin of training, we still apply the ``warmup'' technique that progressively increases the learning rate from $10^{-6}$ to $2\times 10^{-3}$ at the first 5 epochs, then decrease it after 80 and 100 epochs by the factor of 10, and end up at 120.

From Table~\ref{table:coco}, it can be seen that RFB Net300 achieves 30.3\%/49.3\% on the \emph{test-dev} set, which surpasses the baseline score of SSD300* with a large margin, and even equals to that of R-FCN \cite{R-FCN} which employs ResNet-101 as the base net with a larger input size (600$\times$1000) under the two stage framework.

Regarding the bigger model, the result of RFB Net512 is slightly inferior to but still comparable to the one of the recent advanced one-stage model RetinaNet500 (33.8\% vs. 34.4\%). However, it should be noted that RetinaNet makes use of the deep ResNet-101-FPN backbone and a new loss to make learning focus on hard examples, while our RFB Net is only built on a lightweight VGG model. On the other hand, we can see that RFB Net500 averagely consumes 30 ms per image, while RetinaNet needs 90 ms.

One may notice that RetinaNet800 \cite{focal-loss} reports the top accuracy (39.1\%) based on a very high resolution up to 800 pixels. Although it is well known that a larger input image size commonly yields higher performance, it is out of the scope of this study, where an accurate and fast detector is pursued. Instead, we consider two efficient updates: (1) to up-sample the conv7\_fc feature maps and concat it with the conv4\_3 before applying the RFB-s module, sharing a similar strategy as in FPN \cite{FPN}; and (2) to add a branch with a $7\times7$ kernel in all RFB layers. As we can see in Table \ref{table:coco}, they further increase the performance, making the best score in this study at 34.4\% (denoted as RFB Net512-E), while the computational cost only marginally ascends.

\begin{table*}[thbp]
\begin{center}
\scalebox{0.99}{
\resizebox{\textwidth}{!}{
\begin{threeparttable}
\begin{tabular}{c|l|c|ccc|ccc}
\multirow{2}{*}{Method} & \multicolumn{1}{c|}{\multirow{2}{*}{Backbone}}  & \multicolumn{1}{c|}{\multirow{2}{*}{Time}} & \multicolumn{3}{c|}{Avg. Precision, IoU:} & \multicolumn{3}{c}{Avg. Precision, Area:} \\
                        & \multicolumn{1}{c|}{}                           & \multicolumn{1}{c|}{}                      & 0.5:0.95        & 0.5        & 0.75       & S            & M            & L           \\ \Xhline{1.1pt}
Faster \cite{FasterCNN}                  & VGG                                             & 147 ms\                                       & 24.2            & 45.3       & 23.5       & 7.7          & 26.4         & 37.1        \\
Faster+++ \cite{ResNet}               & ResNet-101                                     & 3.36 s\                                      & 34.9            & 55.7       & 37.4       & 15.6         & 38.7         & 50.9        \\
Faster w FPN \cite{FPN}            & ResNet-101-FPN                                 & 240 ms\                                      & 36.2            & 59.1       & 39.0       & 18.2         & 39.0         & 48.2        \\
Faster by G-RMI \cite{trade-off}   & Inception-Resnet-v2 \cite{inceptionv4}           & --                                          & 34.7            &
55.5       & 36.7       & 13.5         &38.1          &52.0        \\
R-FCN \cite{R-FCN}                   & ResNet-101                                    & 110 ms\                                      & 29.9            & 51.9       & --         & 10.8         & 32.8         & 45.0        \\
R-FCN w Deformable CNN \cite{deformableCNN} & ResNet-101                              & 125ms\tnote{\dag}
& 34.5            & 55.0       & --         & 14.0         & 37.7         & 50.3        \\
Mask R-CNN \cite{mask-rcnn}               & ResNext-101-FPN                           & 210 ms\                                      & 37.1            & 60.0       & 39.4       & 16.9         & 39.9         & 53.5        \\ \hline
\hline
YOLOv2 \cite{yolo9000}                  & darknet                                     & 25 ms\                                       & 21.6            & 44.0       & 19.2       & 5.0          & 22.4         & 35.5        \\
SSD300* \cite{ssd}                 & VGG                                            & 12 ms\tnote{\ddag}                                      & 25.1            & 43.1       & 25.8       & --           & --           & --          \\
SSD512* \cite{ssd}                 & VGG                                            & 28 ms\tnote{\ddag}                                      & 28.8            & 48.5       & 30.3       & --           & --           & --          \\
DSSD513 \cite{dssd}                 & ResNet-101                                      & 182 ms\                                      & 33.2            & 53.3       & 35.2       & 13.0         & 35.4         & 51.1        \\
RetinaNet500 \cite{focal-loss}            & ResNet-101-FPN                            & 90 ms                                      & 34.4            & 53.1       & 36.8       & 14.7         & 38.5         & 49.1        \\
RetinaNet800 \cite{focal-loss}            & ResNet-101-FPN                            & 198 ms\                                      & \textbf{39.1} & 59.1       & 42.3       & 21.8         & 42.7         & 50.2        \\ \Xhline{1.1pt}
RFB Net300              & VGG                                                       & \textbf{15 ms}                            & \textbf{30.3}  & 49.3      & 31.8       & 11.8         & 31.9         & 45.9        \\
RFB Net512              & VGG                                                       & \textbf{30 ms}                             & \textbf{33.8}  & 54.2       & 35.9       & 16.2         & 37.1         & 47.4        \\
RFB Net512-E             & VGG                                                 & \textbf{33 ms}
&\textbf{34.4} & 55.7       & 36.4       & 17.6         &37.0          &47.6
\end{tabular}
\begin{tablenotes}
    \footnotesize
    \item[\dag] Extrapolated time
    \item[\ddag] {Tested in Pytorch-0.3.0 and CUDNN V6 for fair comparison}
\end{tablenotes}
\end{threeparttable}}}
\end{center}
\caption{Detection performance on the COCO \emph{test-dev} 2015 dataset. Almost all the methods are measured on the Nvidia Titan X (Maxwell architecture) GPU, except RetinaNet, Mask R-CNN and FPN (Nvidia M40 GPU).}
\label{table:coco}
\end{table*}

%------------------------------------------------------------------------
%------------------------------------------------------------------------
%------------------------------------------------------------------------
\section{Discussion}

\begin{figure}[t]
\centering
\begin{overpic}
[scale=0.6]{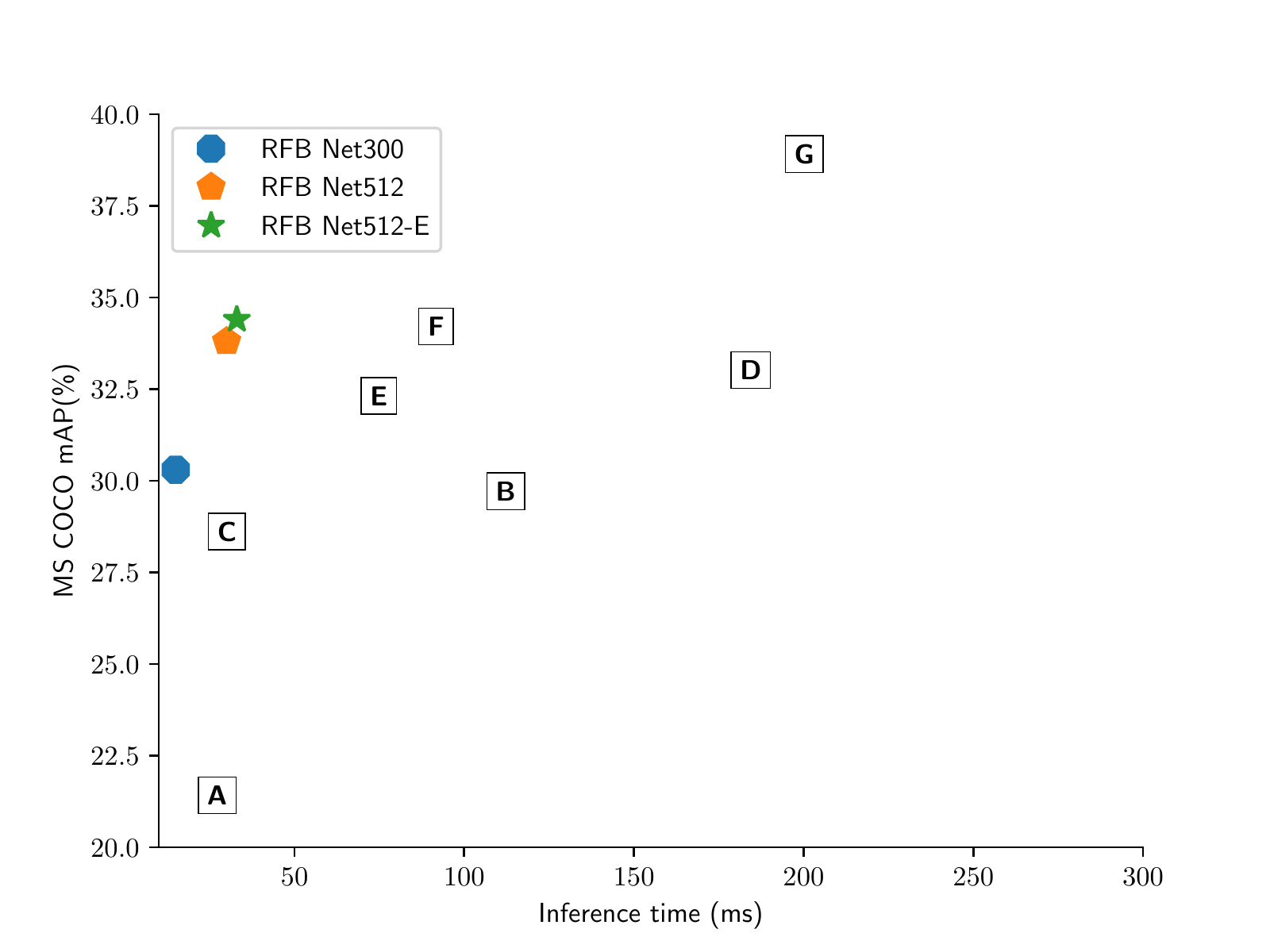}
\put(50,27){
\scalebox{.35}{
\resizebox{\textwidth}{!}{
\begin{tabular}{l|c|c}
                       & mAP(\%)   &  \multicolumn{1}{l}{Time(ms)}  \\ \hline
{[}A{]} YOLOv2 \cite{yolo9000}         & 21.6 & 25   \\
{[}B{]} R-FCN \cite{R-FCN}          & 29.9 & 110  \\
{[}C{]} SSD512* \cite{ssd}        & 28.8 & 28   \\
{[}D{]} DSSD513 \cite{dssd}        & 33.2 & 182  \\
{[}E{]} Retina-50-500 \cite{focal-loss}  & 32.5 & 73   \\
{[}F{]} Retina-101-500 \cite{focal-loss} & 34.4 & 90   \\
{[}G{]} Retina-101-800 \cite{focal-loss} & \textbf{39.1} & 198  \\ \hline
RFB-Net300             & 30.3 & \textbf{15}   \\
RFB-Net512             & 33.8 & 30   \\
RFB-Net512-E           & \textbf{34.4} & 33
\end{tabular}}}}
\end{overpic}
\caption{Speed (ms) vs. accuracy (mAP) on MS COCO \emph{test-dev}. Enabled by the proposed RFB module, our single one-stage detector surpasses all existing high frame rate detectors, including the best reported one-stage system Retina-50-500 \cite{focal-loss}.}
\label{fig:speed}
\end{figure}

\textbf{Inference speed comparison:} In Table~\ref{table:2007} and Fig.~\ref{fig:speed}, we show speed comparison to other recent top-performing detectors. In our experiments, the inference speeds in different datasets have slight variations, since MS COCO has 80 categories and average dense instances consume more time on the NMS process. Table~\ref{table:2007} shows that our RFB Net300 is the most accurate one (80.5\% mAP) among the real-time detectors and runs at 83 fps in Pascal VOC, and RFB Net512 provides more accurate results still with a speed of 38 fps. In Fig.~\ref{fig:speed}, we follow \cite{focal-loss} to plot the speed/accuracy trade-off curve for RFB Net, and compare it to RetinaNet \cite{focal-loss} and other recent methods on the MS COCO \emph{test-dev} set. This plot displays that our RFB Net forms an upper envelope among all the real-time detectors. In particular, RFB Net300 keeps a high speed (66 fps) while outperforming all the high frame rate counterparts. Note that they are measured on the same Titan X (Maxwell architecture) GPU, except RetinaNet (Nvidia M40 GPU).

\textbf{Other lightweight backbone:} Although the base backbone we use is a reduced VGG16 version, it still has a large number of parameters compared with those recent advanced lightweight networks, \emph{e.g.}, MobileNet \cite{mobilenets}, DarkNet \cite{yolo9000}, and ShuffleNet \cite{shufflenet}. To further test the generalization ability of the RFB module, we link RFB to MobileNet-SSD \cite{mobilenets}. Following \cite{mobilenets}, we train it on the MS COCO \emph{train+val35k} dataset with the same schedule and make evaluation on \emph{minival}. Table~\ref{table:mobile} shows that RFB still increases the accuracy of the MobileNet backbone with limited additional layers and parameters. This suggests its great potential for applications on low-end devices.

\textbf{Training from scratch:} We also notice another interesting property of the RFB module, \emph{i.e.} efficiently training the object detector from scratch. Recently, according to \cite{dsod}, training without using pre-trained backbones is discovered to be a hard task, where all the structures of base nets fail to be trained from scratch in the two-stage framework and the prevalent CNNs (ResNet or VGG) in the one-stage framework successfully converge with much worse results. Deeply Supervised Object Detectors (DSOD) \cite{dsod} proposes a lightweight structure which achieves a 77.7\% mAP on the VOC 2007 \emph{test} set without pre-training, but it does not promote the performance when using pre-trained network. We train our RFB Net300 on the VOC 07+12 \emph{trainval} set from scratch and reach a 77.6\% mAP on the same test set, which is comparable to DSOD. It is worth noting that our pre-trained version boosts the performance to 80.5\%.

\begin{table}[thbp]
\centering
\scalebox{.8}{
\resizebox{\textwidth}{!}{
\begin{tabular}{llll}
\Xhline{1.0pt}
\multicolumn{1}{c}{\begin{tabular}[c]{@{}c@{}}Framework\end{tabular}} & \multicolumn{1}{c}{Model} & \multicolumn{1}{c}{mAP (\%)} & \multicolumn{1}{c}{\#parameters} \\ \Xhline{1.0pt}
SSD 300                                                                            & MobileNet \cite{mobilenets}                 & 19.3\%                  & 6.8M                              \\ \hline
SSD 300                                                                            & MobileNet+RFB             & \textbf{20.7\% }        & 7.4M                              \\ \hline
\end{tabular}}}
\caption{Accuracies on MS COCO \emph{minival2014} using MobileNet as the backbone.}
\label{table:mobile}
\end{table}
\section{Conclusion}
In this paper, we propose a fast yet powerful object detector. In contrast to the widely employed way that greatly deepens the backbone, we choose to enhance feature representation of lightweight networks by bringing in a hand-crafted mechanism, namely Receptive Field Block (RFB), which imitates the structure of RF in human visual systems. RFB measures the relationship between the size and eccentricity of RFs, and generates more discriminative and robust features. RFB is equipped on the top of lightweight CNN based SSD, and the resulting detector delivers a significant performance gain on the Pascal VOC and MS COCO databases, where the final accuracies are even comparable to those of existing top-performing deeper model based detectors. In addition, it retains the advantage in processing speed of lightweight models.

\section*{Acknowledgment}
This work was partly supported by the National Key Research and Development Plan (Grant No. 2016YFC0801002) and the National Natural Science Foundation of China (No. 61421003).

\bibliographystyle{splncs04}
\bibliography{eccv}

\end{document}